%% file: main.tex
\documentclass[11pt]{article}

\usepackage[preprint]{acl}

\usepackage{times}
\usepackage{latexsym}

\usepackage[T1]{fontenc}

\usepackage[utf8]{inputenc}

\usepackage{microtype}

\usepackage{inconsolata}

\usepackage{graphicx}

\usepackage{amsmath}
\usepackage{amsfonts}
\usepackage{multirow}
\usepackage{booktabs}
\usepackage{algpseudocode}
\usepackage{tikz}
\usepackage{subcaption}
\usetikzlibrary{positioning}
\usepackage{algorithm}
\usepackage{longtable}
\usepackage{titlesec}
\usepackage{etoolbox}
\usepackage{pifont}

\makeatletter
\patchcmd{\HyField@FlagsRadioButton}{\HyField@SetFlag{Ff}{Radio}}{}{}{}
\makeatother

\title{{TreeFlash: Parallel AR-Approximation for Faster Speculative Decoding}}

\author{Peer Rheinboldt \quad Frédéric Berdoz \quad Roger Wattenhofer \\
  ETH Zurich \\
  \texttt{\{prheinboldt, fberdoz, wattenhofer\}@ethz.ch}}

\begin{document}
\maketitle
\let\oldthefootnote\thefootnote
\renewcommand{\thefootnote}{}
\footnotetext{Code: \url{https://github.com/ETH-DISCO/TreeFlash}}
\renewcommand{\thefootnote}{\oldthefootnote}
\begin{abstract}
  One-shot block drafters for speculative decoding generate the full draft in a single forward pass, achieving strong throughput by eliminating sequential token generation.
  However, they predict each draft token conditioned only on the prefix context, with no dependence on previously drafted tokens.
  This non-autoregressive conditioning causes the drafter's distribution to diverge from the verifier's true autoregressive distribution as draft depth grows. 
  This problem becomes more severe in tree-based drafting, where distinct branches are forced to share the same marginal distribution for subsequent tokens.
  We propose \textbf{TreeFlash}, which addresses this by incorporating an MLP layer conditioned on the drafter's hidden state and the previous token to approximate an autoregressive distribution.
  TreeFlash retains the $\mathcal{O}(1)$ decoding time complexity of one-shot drafters by employing a two-stage approximation mechanism.
  TreeFlash achieves state-of-the-art performance across a variety of tasks and models, improving over marginal tree drafting by $12\%$ higher block efficiency and $9\%$ higher speedup.
\end{abstract}

\section{Introduction}

\begin{figure}[t]
    \centering
    \includegraphics[width=\columnwidth]{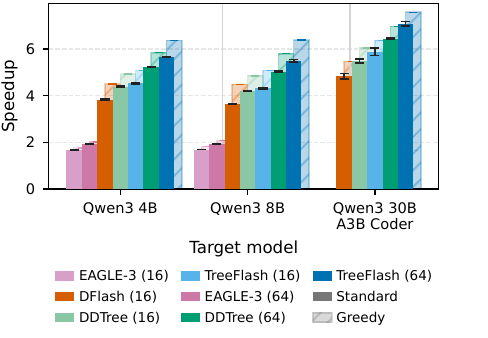}
    \label{fig:q3_8b_speedup_graph}
    \caption{
      Speedup averaged across datasets for standard (solid) and greedy (translucent) decoding.
      Draft budget $B$ is reported in parentheses next to each method.
      TreeFlash consistently outperforms both DFlash ($+17.1\%$) and DDTree ($+3.9\%$) under the same draft budget.
      Under increased draft budget $B=64$ TreeFlash achieves $+9.1\%$ speedup over DDTree.
    }
\end{figure}

\begin{figure*}[!ht]
    \centering
    \includegraphics[width=\textwidth]{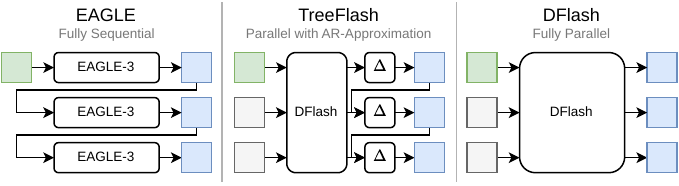}
    \caption{Overview of different drafting paradigms. EAGLE-3 \cite{li2025eagle3} (left) is a small autoregressive drafter. DFlash \cite{chen2026dflash} (right) is a single-pass parallel drafter. TreeFlash (middle) adapts DFlash by including a lightweight AR-approximator layer that allows fully parallel AR-approximation.}
    \label{fig:comparison}
\end{figure*}

The autoregressive nature of transformer-based large language models (LLMs) \cite{vaswani2017attention} fundamentally limits their inference throughput, a constraint that grows increasingly acute as frontier models continue to scale \cite{yan2025qwen3,singh2025openai}.
A paradigm that escapes the quality--efficiency trade-off of compression-based approaches is \emph{speculative decoding} \cite{leviathan2023fast}, in which a lightweight \emph{drafter} proposes a block of $\gamma$ candidate tokens that are then verified in a single parallel forward pass of the larger \emph{verifier}.
This exploits the memory-bandwidth underutilization of standard autoregressive decoding by amortizing costly parameter transfers via batched verification.

A key development in speculative decoding is \emph{tree-based drafting} \cite{miao2024specinfer, sun2023spectr, wang2025opttree}, in which the drafter generates a tree of candidate continuations rather than a single sequence.
By exploring multiple branches simultaneously, tree drafting substantially increases the expected number of accepted tokens per verification step.

A newer frontier in drafting is \emph{one-shot block generation}: rather than producing tokens sequentially, the drafter generates the entire draft block in a single forward pass.
DFlash~\cite{chen2026dflash} shows that a single diffusion-like pass, conditioned on intermediate representations reused from the verifier, achieves substantial speedups over prior autoregressive drafters.
Concurrent to our work, DDTree~\cite{ringel2026ddtree} applies the OPT-Tree construction algorithm of \citet{wang2025opttree} to DFlash's output distribution, yielding further improvements in acceptance and throughput.

However, these one-shot drafters have a fundamental limitation: the predicted distribution for draft token $x_{t+i}$ is conditioned only on the prefix context $x_{\leq t}$, with no dependence on preceding drafted tokens.
This non-autoregressive conditioning causes the drafter's distribution to increasingly diverge from the verifier's true autoregressive distribution as draft depth grows.
In tree-based settings, this problem is compounded: distinct branches that share a common prefix are forced to share the same marginal distribution for subsequent tokens, degrading tree quality.

In light of this limitation, we introduce \textbf{TreeFlash}, a single-pass drafter which incorporates a lightweight AR-approximation mechanism to overcome the conditioning issue while preserving the fully parallel nature of one-shot drafting.

Our key contributions include:
\begin{itemize}
  \item We identify the lack of autoregressive conditioning as a key bottleneck of one-shot block-wise drafters, particularly in tree-based settings.
  \item We introduce a lightweight AR-approximation algorithm that preserves the fully parallel nature of one-shot drafting.
  \item We show that TreeFlash improves the performance over fully marginal distributions and motivate our design choices through ablations of the architecture and training objective.
\end{itemize}

\begin{table*}[!ht]
  \centering
  \def\arraystretch{0.7}
  \setlength{\tabcolsep}{0.1mm}
  \small
  \input{assets/q3_speedup_table}
\caption{%
    Block efficiency ($\tau$) and speedup across datasets and configurations.
    Numbers in \textcolor{gray!65}{grey} denote the tree/draft budget $B$.
    Q3-4B and Q3-8B refer to Qwen3~4B and Qwen3~8B respectively;
    see Table~\ref{tab:q3_30b_results} for Qwen3~Coder~30B~A3B.
    For $T{=}1$ rows, superscripts report standard deviations across run-level means.
    TreeFlash consistently outperforms both baselines across all configurations.
    Under the budget-matched setting ($B{=}16$), transitioning from chain to tree drafting
    (DFlash $\to$ DDTree) already accounts for $64\%$ of the block-efficiency gain
    and $75\%$ of the speedup gain that TreeFlash achieves over DFlash,
    with AR-approximation contributing a further $+0.47$ in $\tau$ and
    $+0.16{\times}$ in speedup on top.
    Under the increased-budget setting ($B{=}64$), the benefit of AR-approximation grows:
    TreeFlash improves over DDTree by $+0.95$ in $\tau$ ($+12.6\%$) and
    $+0.50{\times}$ in speedup ($+9.0\%$).
    We note that EAGLE-3 was
    trained on a different dataset and is included for reference rather than as a direct comparison.
   }
  \label{tab:results}
\end{table*}

\section{Related Work}

\subsection{Speculative Decoding}

Speculative decoding is a lossless paradigm for accelerating autoregressive generation in large language models. 
Early work employs a small, separate language model as a drafter to propose candidate tokens that are subsequently verified by the target model in parallel \cite{leviathan2023fast}.

A key advancement in this line of research is tree-based speculative decoding, in which the drafter constructs a tree of candidate continuations rather than a single sequence, substantially improving acceptance rates. 
Early theoretical treatments derive sophisticated token-acceptance criteria to optimally exploit the drafted tree \cite{miao2024specinfer,sun2023spectr}. 
More recent state-of-the-art methods find that a simple equality-based acceptance check is sufficient in practice, as it admits efficient non-stochastic tree construction strategies \cite{wang2025opttree,li2024eagle2}.

Alongside improvements to tree speculative decoding, considerable effort has been directed at reducing the cost of the drafter itself.
Whereas early work relies on entirely separate small autoregressive models, more recent approaches replace these with lightweight models that reuse intermediate hidden states from the verifier \cite{li2025eagle3}. 
Nevertheless, such models remain fundamentally constrained by autoregressive decoding, limiting the length of the drafted sequence.

To overcome this bottleneck, several works have explored fully parallel draft generation. 
Medusa and Hydra attach linear prediction heads directly to the verifier backbone to predict future tokens simultaneously \cite{cai2024medusa,ankner2024hydra}. 
More recently, diffusion models have been proposed as drafters \cite{li2025diffuspec,sandler2025specdiff2}. 
While diffusion methods eliminate the sequential nature of draft generation, they still require multiple iterations to produce a final draft.

A complementary direction, TiDAR, incorporates parallel drafting directly into the model pretraining objective, enabling a single model to generate draft tokens in parallel and verify them autoregressively, with no observed degradation in output quality relative to standard autoregressive models \cite{liu2025tidar}.

\subsection{One-Shot Block-Wise Drafters}

A recent line of work eliminates multi-step generation entirely, producing the full draft sequence in a single model call. DFlash shows that a single diffusion-like forward pass, conditioned on features reused from the verifier, achieves substantial speedups over prior methods \cite{chen2026dflash}. Concurrent to our work, DDTree extends this by applying the OPT-Tree construction algorithm of \citet{wang2025opttree} to DFlash's output distribution, yielding further improvements in acceptance and throughput \cite{ringel2026ddtree}. DART constructs a tree from independent positional token distributions, but relies on an external N-gram continuity model and the construction of a large N-gram trie at inference time \cite{liu2026dart}.

\section{Method}
\begin{table}[t]
  \def\arraystretch{0.7}
  \setlength\tabcolsep{2pt}
  \small
  \input{assets/q3_30b_table}
  \caption{%
    Block efficiency ($\tau$) and speedup for Qwen3~Coder~30B~A3B on coding benchmarks.
    Notation follows Table~\ref{tab:results}.
    TreeFlash consistently outperforms both DFlash and DDTree across all budgets and temperature settings, confirming that the gains observed on the 4B and 8B target models transfer to a substantially larger MoE architecture.
    At the budget-matched setting ($B{=}16$), TreeFlash improves over DFlash by $+1.50$ in $\tau$ and $+0.99{\times}$ in speedup.
    Under the increased budget ($B{=}64$), $+42.8\%$ speedup over DFlash, with AR-approximation responsible for $+1.01$ block efficiency over DDTree.
  }
  \label{tab:q3_30b_results}
\end{table}

\begin{figure}[t]
    \centering
    \includegraphics[width=\columnwidth]{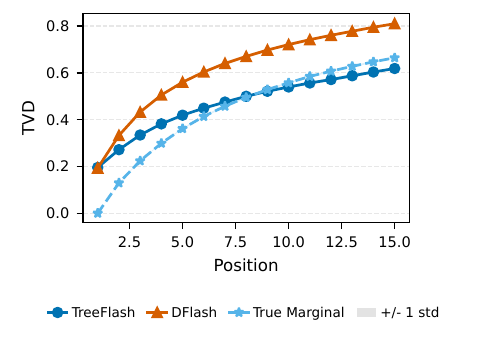}
    \caption{
      Total Variation Distance (TVD) to the verifier distribution across draft positions for Qwen3~4B. 
      We compare DFlash ($q(x_{t+i} | x_{\leq t})$), TreeFlash ($q'(x_{t+i} | x_{\leq t}, x_{t+i-1})$), and an approximation of the ground-truth marginal distribution $p(x_{t+i} | x_{\leq t})$, which is generated using Monte Carlo sampling from the verifier distribution.
      Both DFlash and TreeFlash start with a relatively low TVD of $0.19$ at the first token.
      However, as draft depth increases, DFlash's TVD grows substantially, reaching $0.81$ at depth 15, while TreeFlash's AR-approximation keeps the TVD much lower at $0.62$.
      TreeFlash surpasses the ground-truth marginal distribution at depths beyond 9, suggesting that AR-approximation provides benefits beyond improving the marginal distribution alone.
    }
    \label{fig:tvd_position_graph}
\end{figure}

\paragraph{AR-Approximation.}
A central limitation of block-wise drafters such as DFlash is that they predict only a marginal distribution $q(x_{t+i} | x_{\leq t})$ with no dependence on the immediately preceding tokens.
This is problematic for two reasons.
First, while coherent drafts can still be produced in the greedy ($T{=0}$) single-sequence paradigm, conditioning becomes crucial in the tree-based setting, where multiple parallel candidate sequences are generated.
Second, even under the ground-truth marginal distribution $p(x_{t+i} | x_{\leq t})$, the total variation distance (TVD) to the verifier's AR distribution $p(x_{t+i} | x_{<t+i})$ grows substantially with draft depth, as illustrated in Figure~\ref{fig:tvd_position_graph}.

Some form of conditioning on previously drafted tokens within the block drafter is therefore necessary.
The key challenge is to incorporate such conditioning without sacrificing the $\mathcal{O}(1)$ decoding time complexity that makes one-shot block drafting attractive.
We introduce a lightweight SwiGLU layer that produces a modified hidden state for position $t+i$ by incorporating the input embedding of the preceding token:
\begin{equation}
  h'_{t+i} =  h_{t+i} + \text{SwiGLU}\!\left(\tilde{h}_{t+i} :: e_{t+i-1} \right),
  \label{eq:ar_approx}
\end{equation}
where $\tilde{h}_{t+i}$ is the normalized hidden state and $e_{t+i-1}$ is the normalized input embedding of the preceding token.
We then compute the modified token distribution $q'(x_{t+i} | x_{\leq t}, x_{t+i-1})$ by applying the verifier's output embedding to $h'_{t+i}$.

\paragraph{Inference.}
\label{sec:tree_construction}
We follow the OPT-Tree construction algorithm~\cite{wang2025opttree}, which greedily selects a set of $B$ candidate nodes maximizing the expected number of accepted tokens by retaining those paths with the highest product of drafter token probabilities.
Throughout the paper, $B$ denotes the \emph{draft budget}, i.e., the total number of candidate nodes submitted for verification per speculative decoding step. 
Since token selection is deterministic, we use a simple equality check as the verification procedure, ensuring lossless generation.

Naively applying the AR-approximator of Equation~\ref{eq:ar_approx} requires at least $\gamma$ sequential forward passes through the AR-approximator, partially offsetting the efficiency gains of one-shot drafting.
We avoid this overhead via a two-stage construction.
In the first stage, we use the original DFlash distribution to build a top-$M$-ary tree, where $M$ is a hyperparameter controlling the branching factor (see Section~\ref{abl:inference_parameters} for an empirical analysis).
In the second stage, the AR-approximator is applied to the nodes of this $M$-ary tree to obtain modified token distributions, which are then used to construct the final draft tree via OPT-Tree selection.
Because the AR-approximator only conditions on the preceding token, and because the $M$-ary tree contains exactly $M$ unique tokens at each depth, all modified distributions can be computed in $M \cdot \gamma$ parallel evaluations regardless of $B$, preserving $\mathcal{O}(1)$ drafting time complexity.
Note that this construction requires all non-leaf nodes of the final draft tree to lie within the $M$-ary tree, while leaf nodes may fall outside it.
A detailed description of the algorithm is provided in Appendix~\ref{app:tree_construction}.

\paragraph{Training.}
\label{sec:training}
For each training sample, multiple random anchor positions are sampled at which the drafter gets evaluated.
Ground-truth tokens are used as input for the preceding tokens in the AR-approximation, ensuring the model is trained on faithful input-output pairs.
Diverging from DFlash, we change the loss from cross-entropy on target tokens to forward KL-divergence to the verifier distribution, which has been shown to yield superior draft quality \cite{zhou2024distillspec}.
We initialize the DFlash backbone from the pretrained DFlash checkpoint and zero-initialize the AR-approximator to ensure training starts with a faithful copy of DFlash.
In line with DFlash, we adopt the same loss scaling scheme, penalizing earlier tokens more heavily.

\begin{figure}[t]
    \centering
    \includegraphics[width=\columnwidth]{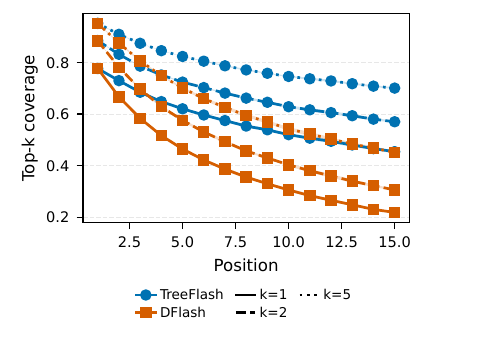}
    \caption{
      Coverage of target probability space with drafter top-$K$ tokens of Qwen3~4B.
      As expected, both DFlash and TreeFlash have the same high coverage at the first token.
      With a budget of just 2 tokens, both achieve a coverage of $0.88$ for the first token.
      However, the further the token is in the future, the more pronounced the benefit of AR-approximation becomes.
      Already for the third token, TreeFlash's top-$2$ coverage is $0.79$ compared to DFlash's $0.69$.
      At depth 15, TreeFlash's top-$1$ coverage is $0.45$, which is slightly larger than DFlash's top-$5$ coverage of $0.44$.
    }
    \label{fig:topk_coverage}
\end{figure}

\section{Experiments}
\label{sec:experiments}
\paragraph{Baselines. }
We evaluate against three state-of-the-art drafters.
\emph{EAGLE-3}~\cite{li2025eagle3} is a small autoregressive drafter that generates tokens sequentially.
We use the checkpoints provided by AngelSlim~\cite{AngelSlim2025}. It is important to note that these were trained on a different data distribution from DFlash and TreeFlash. 
\emph{DFlash}~\cite{chen2026dflash} produces a sequence draft in a single forward pass, and
\emph{DDTree}~\cite{ringel2026ddtree} applies the OPT-Tree construction algorithm on top of this.
DFlash, DDTree, and TreeFlash share the same underlying pretrained checkpoint.
TreeFlash additionally fine-tunes this checkpoint as described in Section~\ref{sec:training} and adds extra parameters; see Table~\ref{tab:ablations} for an equivalently finetuned DFlash model.
Target models are Qwen3~4B~($+125M$), Qwen3~8B~($+251M$), and Qwen3~Coder~30B~A3B~($+63M$) \cite{yan2025qwen3}.

\paragraph{Datasets. }
We evaluate across a diverse set of tasks.
For mathematical reasoning, we use \emph{MATH-500}~\cite{lightman2024lets} and \emph{GSM8K}~\cite{cobbe2021training}.
Code generation is assessed on \emph{HumanEval}~\cite{chen2021evaluating} and \emph{MBPP}~\cite{austin2021program}.
General instruction-following ability is evaluated using \emph{MT-Bench}~\cite{zheng2023judging}.
Each dataset is limited to 64 samples.

\paragraph{Training Setup. }
Taking inspiration from DFlash \cite{chen2026dflash}, we train on a 100k-sample subset of a synthetic dataset generated from the Nemotron Post-Training Dataset V2~\cite{nathawani2025nemotronposttraining} and CodeAlpaca~\cite{chaudhary2023codealpaca} prompts.
All models are trained for one epoch with an effective batch size of 128 and 128 anchors per sample.
We use a linear learning rate warmup for 128 steps, followed by a cosine decay schedule.
Optimization is performed using AdamW with a peak learning rate of $10^{-4}$ \cite{loshchilovdecoupled}. 
Loss scaling uses a factor of $7$ as in DFlash, and sequences are limited to 3072 tokens.

\paragraph{Metrics. }
In speculative decoding, the main metric of interest is \emph{speedup}, which measures the increase in throughput compared to vanilla decoding.
This metric, however, depends on multiple factors, such as implementation and hardware.
Therefore, a commonly used metric for measuring drafter quality is \emph{block efficiency ($\tau$)}, which measures the number of accepted tokens plus the additional residual token generated from the verification step.
This is equivalent to the average number of tokens generated per draft-verify iteration.
Furthermore, to measure drafter quality, we report the \emph{total variation distance (TVD)} compared to the verifier's autoregressive distribution $p(x_t | x_{<t})$.
Furthermore, we report \emph{top-$K$ coverage}, which measures the cumulative probability of the tokens in the drafter's top-$K$ set under the verifier distribution.
See Appendix~\ref{app:tvd_approx} for details on the calculation of TVD and top-$K$ coverage.

\paragraph{Inference Parameters. }
We use a block size of $\gamma=16$, and for TreeFlash, we use $M=16$ for the tree construction.
Drafter temperatures are set to $T=1$ regardless of the verifier temperature.
We limit the maximum output length to 2048 additional tokens for all samples.
We use PyTorch SDPA as the attention implementation during inference and BFloat16 precision.
Experiments are conducted on NVIDIA~GH200 GPUs.

\begin{table}[t]
  \def\arraystretch{0.7}
  \setlength\tabcolsep{2pt}
  \small
  \input{assets/abl_model_table}
  \caption{
    Block efficiency for the ablation experiments with $B=64$. 
    Results are on 64 samples for \emph{MATH-500}, \emph{HumanEval}, and \emph{MT-Bench}.
    \emph{w/o AR-app}roximation is an extended finetune using the DFlash paradigm, 
    \emph{w/ Linear} replaces the SwiGLU layer with a linear layer,
    \emph{w/ Frozen} keeps the DFlash backbone frozen during training,
    \emph{w/ 2-prev} inputs the two previous tokens to the AR-approximation, 
    \emph{w/ CE} replaces the KL loss with cross-entropy, and
    \emph{w/o Scaling} removes the loss scaling scheme from DFlash training. 
    Using a SwiGLU layer instead of a linear layer provides the largest improvement in block efficiency.
    Further, keeping the backbone frozen has similar performance to full TreeFlash, indicating that the majority of the gains are attributable to the AR-approximator itself rather than continued backbone fine-tuning.
   }
  \label{tab:ablations}
\end{table}
\subsection{Results}
\paragraph{Speculative Decoding. }

\begin{figure*}[t]
  \centering

  \begin{minipage}[t]{8.5cm}
    \centering
    \includegraphics[width=\linewidth]{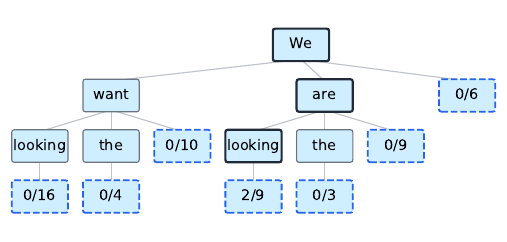}

    \vspace{0.25em}
    \emph{(i) DDTree}
  \end{minipage}
  \hfill
  \begin{minipage}[t]{7cm}
    \centering
    \includegraphics[width=\linewidth]{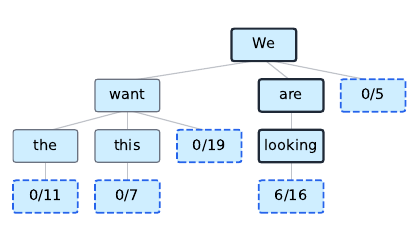}

    \vspace{0.25em}
    \emph{(ii) TreeFlash}
  \end{minipage}

  \caption{
    Qualitative comparison of decoding sub-trees produced by DDTree and TreeFlash.
    Outline tokens indicate acceptance, and $i/j$ denotes $i$ nodes accepted from $j$ total. 
    Note that TreeFlash produces more coherent bigrams and isn't limited to marginal distributions, allowing for better utilization of draft budget. See Appendix~\ref{app:full_qual} for the full trees.
  }
  \label{fig:treeflash_qualitative}
\end{figure*}

Tables~\ref{tab:results} and~\ref{tab:q3_30b_results} report speedup and block efficiency across all targets, tasks, and decoding regimes. 
Figure~\ref{fig:treeflash_qualitative} shows a qualitative example of a drafted tree comparing DDTree's marginally guided construction with TreeFlash's AR-approximation guided draft tree.
TreeFlash consistently outperforms all baselines in every evaluated configuration.
As can be seen in Tables~\ref{tab:results} and~\ref{tab:q3_30b_results}, under a budget-matched setting ($B{=}16$), TreeFlash improves over DFlash by an average of $+1.35$ in $\tau$ ($+24.8\%$) and $+0.69{\times}$ in speedup ($+17.1\%$).

The major driver of this improvement is the transition from chain to tree drafting.
As can be seen in Tables~\ref{tab:results} and~\ref{tab:q3_30b_results}, DDTree alone recovers ${\sim}63\%$ of the block-efficiency gain and ${\sim}72\%$ of the speedup gain that TreeFlash achieves over DFlash.
Nevertheless, AR-approximation contributes a consistent and meaningful improvement on top:
as can be seen in Tables~\ref{tab:results} and~\ref{tab:q3_30b_results}, TreeFlash outperforms DDTree by $+0.50$ in $\tau$ ($+7.5\%$) and $+0.19{\times}$ in speedup ($+3.9\%$) under matched budget.

A key advantage of tree drafting is that it enables scaling $B$ beyond the block size.
As can be seen in Tables~\ref{tab:results} and~\ref{tab:q3_30b_results}, under the increased budget of $B{=}64$, DDTree already substantially improves over DFlash by $+34.9\%$ $\tau$ and $+34.1\%$ speedup on average.
TreeFlash pushes this further to $+51.6\%$ and $+46.2\%$ respectively.

Notably, the benefit of AR-approximation grows with draft budget $B$.
As can be seen in Tables~\ref{tab:results} and~\ref{tab:q3_30b_results}, at $B{=}16$ TreeFlash improves over DDTree by $+7.5\%$ in $\tau$ and $+3.9\%$ in speedup, whereas at $B{=}64$ these gains increase to $+12.4\%$ and $+9.1\%$ respectively.
We attribute this to the fact that larger budgets expose more tokens at deeper draft positions. 
As shown in Figures~\ref{fig:tvd_position_graph} and~\ref{fig:topk_coverage}, this is where AR-approximation provides the greatest improvement over marginal distributions.
In line with this, the benefit of AR-approximation is most pronounced in high-acceptance regimes.
As can be seen in Table~\ref{tab:results}, on MATH-500 and $B=64$ TreeFlash improves $\tau$ over DDTree by $+13.3\%$, whereas on MT-Bench the gain is only $+10.2\%$.

In summary, TreeFlash consistently and meaningfully outperforms both DFlash and DDTree across all evaluated models, tasks, and decoding regimes.
While the transition to tree-based drafting provides the largest single boost in performance, AR-approximation contributes a consistent, complementary gain that grows with draft budget and is most pronounced in high-acceptance settings.

\begin{figure}[t]
    \centering
    \includegraphics[width=\columnwidth]{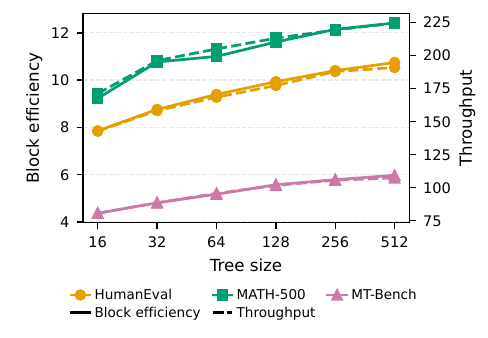}
    \caption{%
      Block efficiency and throughput for different draft budgets $B$ for TreeFlash with Qwen3~4B as the target model under greedy decoding.
      Increasing the tree budget consistently improves both metrics, with mean block efficiency growing from $7.15$ at $B{=}16$ to $9.71$ at $512$.
      Note that while in single-batch settings the draft budget can be scaled to large values, this is not the case in multi-batch settings, where the compute overhead of tree verification typically favors smaller draft budgets.
    }
    \label{fig:treesize_graph}
\end{figure}

\begin{figure}
    \centering
    \includegraphics[width=\columnwidth]{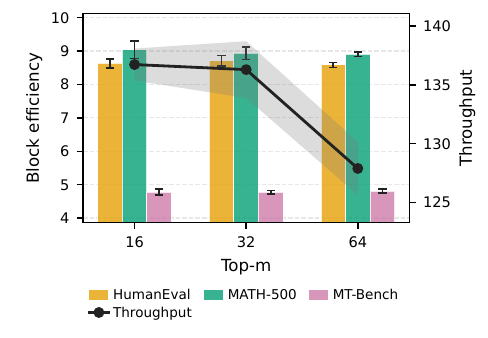}
    \caption{%
      Block efficiency and throughput for different values of $M$ for TreeFlash with Qwen3~4B as the target model under greedy decoding with $B{=}64$.
      In terms of block efficiency, the values are similar; however, for throughput, the cutoff appears to be around $M=32$, where larger values incur too much overhead and cause throughput to drop.
      This value depends on the underlying hardware and implementation.
    }
    \label{fig:top_m_graph}
\end{figure}

\paragraph{Autoregressive Approximation. }
TVD to the target distribution serves as a useful diagnostic for the calibration between drafter and verifier.

As shown in Figure~\ref{fig:tvd_position_graph}, the TVD of DFlash grows steadily with draft depth, starting at $0.19$ for the first drafted token and reaching $0.81$ by the end of the block.
Notably, this degradation is not specific to DFlash but reflects a systematic limitation of marginal distributions. As can be seen, the true marginal $p(x_{t+i} \mid x_{\leq t})$ (see Appendix~\ref{app:tree_construction}) also exhibits increasing divergence with depth.
TreeFlash is not immune to this effect, but its growth in TVD is substantially slower, topping out at $0.62$.
Remarkably, beyond the 9th draft position the true marginal itself exhibits higher TVD than TreeFlash, suggesting that conditioning on the preceding token yields a better calibrated distribution compared to the ground-truth marginal.

Figure~\ref{fig:topk_coverage} shows top-$K$ coverage, which correlates the number of alternative tokens in the draft tree to the acceptance probability. 
Again, for tokens at depth $1$, TreeFlash and DFlash both achieve a top-$1$ coverage of $0.78$.
As depth increases, however, the gap widens considerably: at depth 10 and beyond, TreeFlash's top-$1$ coverage approximately matches DFlash's top-$5$ coverage.
Concretely, this means that a single TreeFlash candidate achieves the same acceptance probability as $5$ DFlash candidates, allowing for more efficient token utilization at depth.

In conclusion, the AR-approximation's benefits can be observed not only in empirical speculative decoding measures but also in distributional metrics.
We show that TreeFlash's distribution is not only better than DFlash's, but also better than the true marginal distribution, suggesting that the gained information from conditioning on the preceding token captures token-level coherence that marginal distributions systematically fail to exploit.

\subsection{Ablations}

\paragraph{Ablation Setup. }
All ablation experiments are conducted on Qwen3~4B. The training setup is equivalent to the main experiments, unless stated otherwise. Evaluation is performed on 64 samples of MATH-500, HumanEval, and MT-Bench with a tree size of 64.

\paragraph{Model Design. }
The design of the AR-approximator is critical to the efficacy of TreeFlash: it must be both lightweight enough to preserve the $\mathcal{O}(1)$ drafting complexity and expressive enough to meaningfully correct the hidden states.
First, we finetune the DFlash model with no AR-approximation mechanism, which, as can be seen in Table~\ref{tab:ablations}, fails to improve over the initial checkpoint.
As an alternative to the SwiGLU head, we evaluate a simple linear map that applies a bias to the hidden state conditioned on the previous token.
As shown in Table~\ref{tab:ablations}, this linear approximator consistently underperforms TreeFlash in block efficiency across all evaluated settings, confirming that the non-linear capacity of the SwiGLU head is necessary.

We further show that a single previous token is sufficient for AR-approximation.
As shown in Table~\ref{tab:ablations}, conditioning on the previous bigram (w/ 2-prev) provides little to no improvement over TreeFlash. 
Given the increase in complexity and growth of evaluations required, we conclude that single-token conditioning strikes the best balance between performance and efficiency.

\paragraph{Training Paradigm. }
TreeFlash jointly fine-tunes both the AR-approximator and the DFlash backbone.
To isolate the contribution of the AR-approximator, we evaluate a frozen-backbone variant (\emph{w/ Frozen}), which keeps the DFlash checkpoint fixed and trains only the SwiGLU head.
As shown in Table~\ref{tab:ablations}, this variant performs only marginally worse than full TreeFlash, indicating that the majority of the gains are attributable to the AR-approximator itself rather than continued backbone fine-tuning.

Diverging from DFlash, TreeFlash uses forward KL divergence rather than cross-entropy to align the drafter to the target distribution. 
Table~\ref{tab:ablations} shows that replacing the KL divergence with cross-entropy (\emph{w/ CE}) yields comparable performance, with TreeFlash retaining a slight edge.
Like DFlash, TreeFlash applies a loss scaling mechanism that penalizes early errors more heavily than later errors.
Removing this scaling (\emph{w/o Scaling}) produces mixed results: block efficiency is marginally higher without scaling on long-acceptance tasks such as MATH-500 and code, while scaling provides a small benefit in low-acceptance regimes such as MT-Bench.

Overall, the training design choices each contribute modest improvements, showing that their impact is secondary to the architectural choices analyzed above.

\paragraph{Inference Parameters. }
\label{abl:inference_parameters}
The two primary inference-time hyperparameters of TreeFlash are the drafter token budget $B$ and the candidate set size $M$.

The token budget $B$ controls the node count of the drafted tree, and larger values are expected to increase block efficiency $\tau$ at the cost of additional verification compute.
As shown in Figure~\ref{fig:treesize_graph}, $B$ can be scaled to large values without the verification overhead outweighing the gains in block efficiency.
In practice, the optimal $B$ depends on multiple factors, such as batch size, hardware, and memory constraints, and should be tuned accordingly.

The hyperparameter $M$ controls the number of candidate tokens considered per position during tree construction, and directly determines the number of evaluations required by the AR-approximator.
As shown in Figure~\ref{fig:top_m_graph}, block efficiency is comparable across different $M$ values, while throughput favors $M \leq 32$.
Therefore, we recommend choosing $M \leq 32$, as these values provide the best throughput in our experiments.

\section{Conclusion}
This paper presents TreeFlash, a method to improve one-shot block-wise drafting by incorporating a lightweight AR-approximation mechanism that conditions each draft position on the preceding token, preserving the $\mathcal{O}(1)$ decoding complexity of parallel drafters while addressing the distributional gap that grows with draft depth.
TreeFlash achieves consistent and substantial performance improvements over both DFlash and DDTree across a wide range of tasks, model sizes, and decoding regimes, improving block efficiency by $+12.4\%$ over trees constructed with fully marginal distributions.

\section*{Limitations}
\label{sec:limitations}
TreeFlash demonstrates consistent improvements across all evaluated tasks, model sizes, and decoding regimes.

All experiments are conducted in a single-batch setting using SDPA attention.
Production serving systems, however, typically incorporate a range of optimizations, such as quantization, multi-batch decoding, and custom attention kernels, that can conflict with large tree sizes and dynamic tree topologies~\cite{dao2022flashattention, lin2024awq, kwon2023pagedattention}.
The interaction between TreeFlash and such serving-level optimizations is not evaluated in this work.

TreeFlash is initialised from an existing DFlash checkpoint rather than trained from scratch.
A full pretraining run is beyond the scope of this paper, and would likely require an auxiliary loss on the initial candidate distribution used to construct the $M$-ary tree.
The same concern applies to extended training of the AR-approximator: prolonged fine-tuning of the DFlash backbone may degrade the initial candidate distributions used to construct the $M$-ary tree, leading to suboptimal final tree construction

The AR-approximator is trained with teacher-forced input embeddings, which is the standard paradigm for autoregressive models but may be suboptimal here: at inference time, the approximator conditions on its own previously drafted tokens rather than ground-truth ones.
Addressing this exposure bias is left to future work.
Recent work on training models explicitly for tree generation rather than sequential AR modelling may also offer a promising future direction in training TreeFlash~\cite{hu2026bridging}.

TreeFlash is evaluated exclusively on Qwen-family models.
Whether the efficiency gains transfer to other model families remains an open question.

\bibliography{references}
\newpage
\appendix
\label{sec:appendix}
\section{TVD and Coverage approximation}
\label{app:tvd_approx}
To compute the target marginal distribution $p(x_{t+i} \mid x_{\leq t})$, we employ Monte Carlo estimation.
For a given anchor position $t$, we draw $N$ independent continuations $x^{(j)}_{t+1:t+\gamma}$, $j \in [N]$, and approximate the marginal as
\begin{equation}
    \tilde{p}(x_{t+i} \mid x_{\leq t}) \approx \frac{1}{N} \sum_{j=1}^{N} p\!\left(x_{t+i} \mid x^{(j)}_{<t+i}\right).
\end{equation}
, where $x^{(j)}_{<t+i-1} := x_{\leq t}, x^{(j)}_{t+1:t+i-1} $
Note that this is an unbiased estimator of the true marginal, i.e.,  $\mathbb{E}\left[ \tilde{p}(x_{t+i} \mid x_{\leq t})\right] = p(x_{t+i} \mid x_{\leq t})$.
The TVD between the target marginal and the target distribution is approximated with:
\begin{equation}
    \frac{1}{N} \sum_{j=1}^{N} TVD\left(\tilde{p}(x_{t+i} \mid x_{\leq t}),\   p\!\left(x_{t+i} \mid x^{(j)}_{<t+i}\right) \right)
\end{equation}
Note that while not an unbiased estimator, this serves as a lower bound to the ground truth TVD by Jensen's Inequality. 

Top-$K$ coverage is defined as follows: let $\hat{x}^{(1)}_{t+i}, \ldots, \hat{x}^{(k)}_{t+i}$
denote the top-$K$ tokens according to the drafter; coverage at depth $i$ is then defined as
\begin{equation}
    C_k = \sum_{l=1}^{k} p\!\left(x_{t+i} = \hat{x}^{(l)}_{t+i} \mid x_{<t+i}\right).
\end{equation}
For all evaluations, we sample 512 examples from the held-out validation set, drawing
$16$ anchor positions per sample with $N{=}64$ continuations each.
To limit memory consumption, autoregressive probabilities are truncated to the top-$256$
tokens.
All results are averaged over three random seeds, with standard deviations reported in the
figures.

\section{Tree Construction}
\label{app:tree_construction}
Figure~\ref{alg:ar_opttree_construction} is the algorithm used for efficient tree construction. 
Note that Part~1 can be done efficiently in parallel on the GPU and Part~2 is done iteratively on the CPU.

\begin{figure}
  \include{assets/tree_construction}
  \caption{AR-approximated OPT-Tree Construction. $\gamma$ is the maximum draft depth, $M$ is the hyperparameter controlling the size of the initial tree, and $B$ is the tree/draft budget.}
  \label{alg:ar_opttree_construction}
\end{figure}

\section{Qualitative Trees}
\label{app:full_qual}
\begin{figure*}
    \includegraphics[]{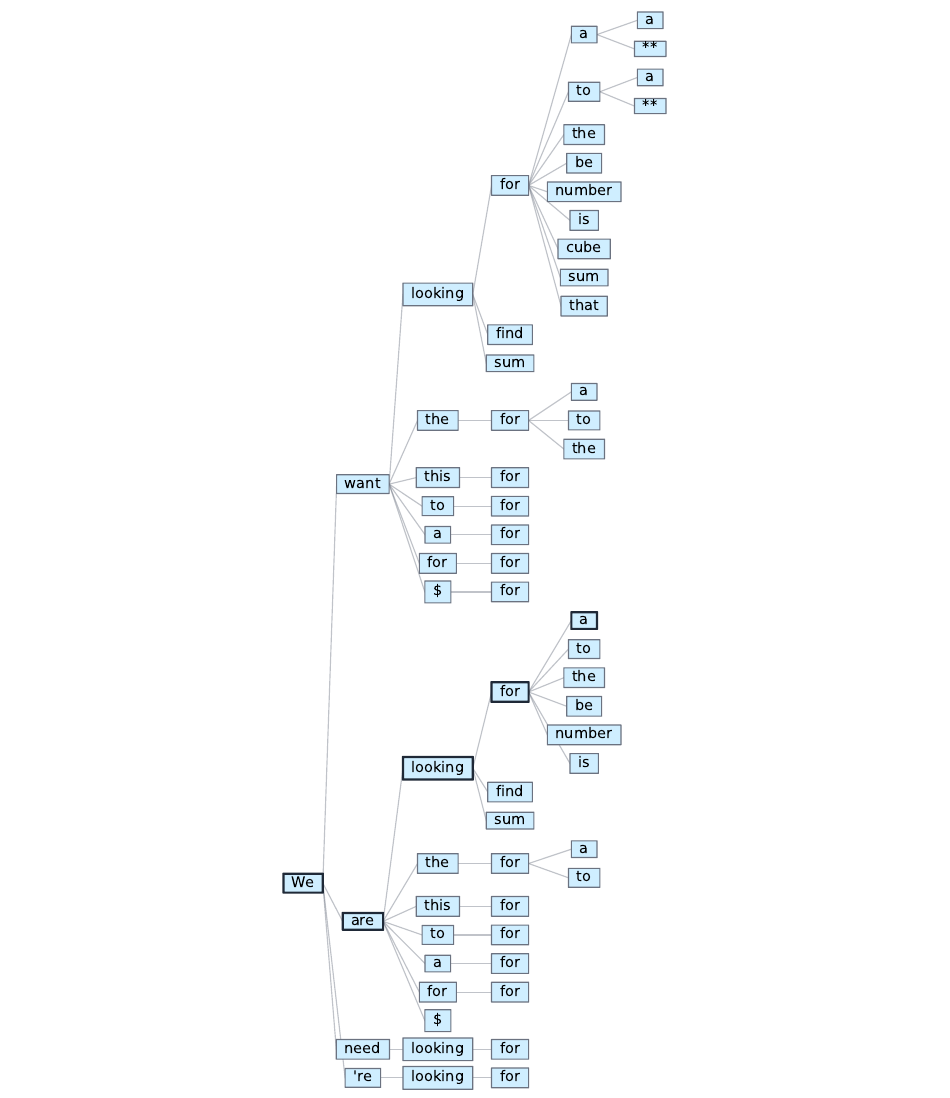}
    \caption{Example of a Draft Tree with $B{=}64$ produced by DDTree. Outlined tokens are accepted by Qwen3~8B under greedy sampling.}
    \label{fig:qual_ddtree_full}
\end{figure*}

\begin{figure*}
    \includegraphics[]{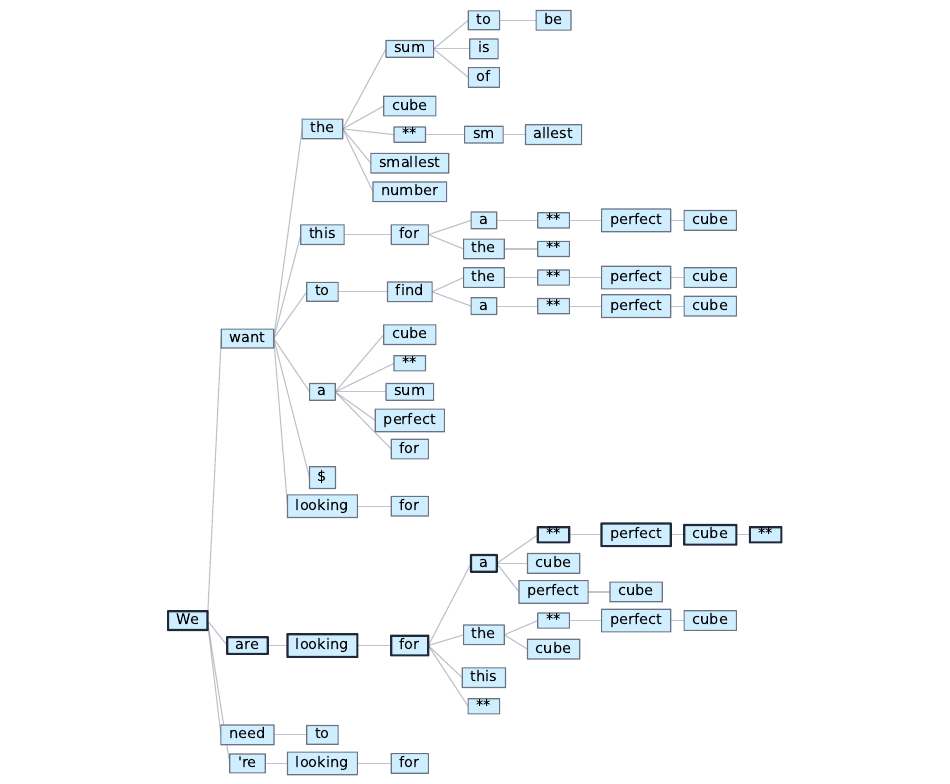}
    \caption{Example of a Draft Tree with $B{=}64$ produced by TreeFlash. Outlined tokens are accepted by Qwen3~8B under greedy sampling.}
    \label{fig:qual_treeflash_full}
\end{figure*}

As illustrated in Figures~\ref{fig:qual_ddtree_full} and~\ref{fig:qual_treeflash_full}, the two methods produce qualitatively different tree topologies.
Because DDTree relies solely on the marginal distribution, the token ranking at each depth is identical across all paths.
This forces DDTree to always produce a \emph{nested} tree: for any two sibling nodes $i$ and $j$ at depth $d$, if $i$ has a higher marginal probability than $j$, then the set of descendants of $j$ must be a subtree of the descendants of $i$.
Put differently, siblings at depth $d$ are constrained to share the same child ordering, since no path-specific information is available to differentiate them.
As shown in Figure~\ref{fig:qual_treeflash_full}, TreeFlash escapes this constraint:
By conditioning on the draft context via the AR-approximator, different paths can yield different child rankings, allowing the tree to better reflect the true conditional structure of the continuations and resulting in better-calibrated topologies.

\section{Licenses}
\label{app:licenses}
All Qwen3 model checkpoints, MT-Bench, and CodeAlpaca are released under the Apache~2.0 license. 
MATH-500, GSM8K, HumanEval, and the DFlash checkpoint are released under the MIT license. 
MBPP is released under CC-BY-4.0; the Nemotron Post-Training Dataset~V2 is predominantly released under CC-BY-4.0, with small subsets under ODC-BY and CC-BY-SA.


\end{document}

%% file: assets/q3_speedup_table.tex
\begin{tabular*}{\textwidth}{@{\extracolsep{\fill}}l*{12}{c}}
\toprule
\multirow{2}{*}{Method} & \multicolumn{2}{c}{GSM8K} & \multicolumn{2}{c}{MATH-500} & \multicolumn{2}{c}{HumanEval} & \multicolumn{2}{c}{MBPP} & \multicolumn{2}{c}{MT-Bench} & \multicolumn{2}{c}{Mean} \\
 & Speedup & $\tau$ & Speedup & $\tau$ & Speedup & $\tau$ & Speedup & $\tau$ & Speedup & $\tau$ & Speedup & $\tau$ \\
\midrule
\multicolumn{13}{c}{T=1 (Sampling)} \\
\midrule
\addlinespace[1pt]
\multicolumn{13}{@{}l}{\textbf{Target model: Qwen3 4B}} \\
\addlinespace[1pt]
EAGLE3 \textcolor{gray!65}{(16)} & 1.79$^{0.01}$ & 3.15$^{0.01}$ & 1.68$^{0.01}$ & 2.93$^{0.01}$ & 1.70$^{0.01}$ & 2.96$^{0.01}$ & 1.64$^{0.02}$ & 2.88$^{0.02}$ & 1.62$^{0.02}$ & 2.84$^{0.02}$ & 1.69$^{0.01}$ & 2.95$^{0.01}$ \\
DFlash \textcolor{gray!65}{(16)} & 4.19$^{0.06}$ & 5.65$^{0.08}$ & 4.39$^{0.09}$ & 5.87$^{0.12}$ & 4.24$^{0.05}$ & 5.66$^{0.05}$ & 3.96$^{0.08}$ & 5.31$^{0.08}$ & 2.34$^{0.02}$ & 3.08$^{0.02}$ & 3.82$^{0.04}$ & 5.11$^{0.04}$ \\
DDTree \textcolor{gray!65}{(16)} & 4.78$^{0.08}$ & 6.66$^{0.10}$ & 4.96$^{0.08}$ & 6.86$^{0.08}$ & 4.82$^{0.07}$ & 6.68$^{0.11}$ & 4.52$^{0.10}$ & 6.22$^{0.17}$ & 2.84$^{0.01}$ & 3.89$^{0.01}$ & 4.38$^{0.03}$ & 6.06$^{0.05}$ \\
TreeFlash \textcolor{gray!65}{(16)} & \textbf{4.98$^{0.04}$} & \textbf{7.15$^{0.07}$} & \textbf{5.12$^{0.13}$} & \textbf{7.31$^{0.19}$} & \textbf{4.99$^{0.08}$} & \textbf{7.13$^{0.08}$} & \textbf{4.67$^{0.07}$} & \textbf{6.73$^{0.09}$} & \textbf{2.88$^{0.01}$} & \textbf{4.08$^{0.01}$} & \textbf{4.53$^{0.03}$} & \textbf{6.48$^{0.04}$} \\
\cmidrule(lr){1-13}
EAGLE3 \textcolor{gray!65}{(64)} & 2.06$^{0.01}$ & 3.66$^{0.03}$ & 1.91$^{0.02}$ & 3.41$^{0.01}$ & 1.91$^{0.02}$ & 3.39$^{0.01}$ & 1.87$^{0.01}$ & 3.32$^{0.02}$ & 1.83$^{0.00}$ & 3.27$^{0.00}$ & 1.92$^{0.01}$ & 3.41$^{0.01}$ \\
DDTree \textcolor{gray!65}{(64)} & 5.71$^{0.05}$ & 7.74$^{0.06}$ & 5.88$^{0.08}$ & 7.92$^{0.09}$ & 5.81$^{0.06}$ & 7.82$^{0.05}$ & 5.39$^{0.03}$ & 7.29$^{0.01}$ & 3.36$^{0.01}$ & 4.47$^{0.00}$ & 5.23$^{0.02}$ & 7.05$^{0.03}$ \\
TreeFlash \textcolor{gray!65}{(64)} & \textbf{6.22$^{0.01}$} & \textbf{8.72$^{0.04}$} & \textbf{6.34$^{0.09}$} & \textbf{8.84$^{0.14}$} & \textbf{6.23$^{0.15}$} & \textbf{8.67$^{0.21}$} & \textbf{5.95$^{0.11}$} & \textbf{8.34$^{0.12}$} & \textbf{3.59$^{0.01}$} & \textbf{4.92$^{0.01}$} & \textbf{5.67$^{0.01}$} & \textbf{7.90$^{0.02}$} \\
\midrule
\addlinespace[1pt]
\multicolumn{13}{@{}l}{\textbf{Target model: Qwen3 8B}} \\
\addlinespace[1pt]
EAGLE3 \textcolor{gray!65}{(16)} & 1.82$^{0.01}$ & 3.13$^{0.01}$ & 1.72$^{0.00}$ & 2.94$^{0.01}$ & 1.78$^{0.01}$ & 3.06$^{0.02}$ & 1.63$^{0.00}$ & 2.80$^{0.01}$ & 1.51$^{0.01}$ & 2.58$^{0.01}$ & 1.69$^{0.00}$ & 2.90$^{0.00}$ \\
DFlash \textcolor{gray!65}{(16)} & 4.21$^{0.06}$ & 5.65$^{0.07}$ & 4.26$^{0.06}$ & 5.68$^{0.05}$ & 3.95$^{0.05}$ & 5.26$^{0.06}$ & 3.64$^{0.04}$ & 4.93$^{0.04}$ & 2.19$^{0.01}$ & 2.89$^{0.00}$ & 3.65$^{0.01}$ & 4.88$^{0.02}$ \\
DDTree \textcolor{gray!65}{(16)} & 4.77$^{0.03}$ & 6.62$^{0.03}$ & 4.84$^{0.09}$ & 6.66$^{0.11}$ & 4.46$^{0.07}$ & 6.16$^{0.09}$ & 4.25$^{0.03}$ & 5.91$^{0.03}$ & 2.68$^{0.04}$ & 3.65$^{0.04}$ & 4.20$^{0.02}$ & 5.80$^{0.02}$ \\
TreeFlash \textcolor{gray!65}{(16)} & \textbf{4.85$^{0.03}$} & \textbf{7.03$^{0.03}$} & \textbf{4.88$^{0.07}$} & \textbf{7.02$^{0.06}$} & \textbf{4.63$^{0.03}$} & \textbf{6.63$^{0.04}$} & \textbf{4.44$^{0.03}$} & \textbf{6.37$^{0.05}$} & \textbf{2.76$^{0.00}$} & \textbf{3.92$^{0.01}$} & \textbf{4.31$^{0.01}$} & \textbf{6.19$^{0.01}$} \\
\cmidrule(lr){1-13}
EAGLE3 \textcolor{gray!65}{(64)} & 2.06$^{0.01}$ & 3.63$^{0.01}$ & 1.94$^{0.00}$ & 3.41$^{0.02}$ & 2.03$^{0.01}$ & 3.55$^{0.01}$ & 1.86$^{0.02}$ & 3.25$^{0.04}$ & 1.71$^{0.01}$ & 2.98$^{0.01}$ & 1.92$^{0.00}$ & 3.36$^{0.01}$ \\
DDTree \textcolor{gray!65}{(64)} & 5.67$^{0.04}$ & 7.67$^{0.07}$ & 5.73$^{0.09}$ & 7.65$^{0.12}$ & 5.40$^{0.10}$ & 7.26$^{0.12}$ & 5.15$^{0.11}$ & 6.95$^{0.13}$ & 3.22$^{0.05}$ & 4.28$^{0.05}$ & 5.04$^{0.03}$ & 6.76$^{0.03}$ \\
TreeFlash \textcolor{gray!65}{(64)} & \textbf{6.21$^{0.12}$} & \textbf{8.69$^{0.18}$} & \textbf{6.30$^{0.05}$} & \textbf{8.70$^{0.04}$} & \textbf{5.88$^{0.09}$} & \textbf{8.13$^{0.11}$} & \textbf{5.66$^{0.06}$} & \textbf{7.86$^{0.07}$} & \textbf{3.46$^{0.02}$} & \textbf{4.70$^{0.03}$} & \textbf{5.50$^{0.05}$} & \textbf{7.62$^{0.06}$} \\
\midrule
\multicolumn{13}{c}{T=0 (Greedy)} \\
\midrule
\addlinespace[1pt]
\multicolumn{13}{@{}l}{\textbf{Target model: Qwen3 4B}} \\
\addlinespace[1pt]
EAGLE3 \textcolor{gray!65}{(16)} & 1.84 & 3.18 & 1.78 & 3.09 & 1.75 & 3.05 & 1.69 & 2.94 & 1.71 & 2.97 & 1.75 & 3.05 \\
DFlash \textcolor{gray!65}{(16)} & 4.63 & 6.26 & 5.96 & 7.93 & 4.87 & 6.44 & 4.37 & 5.83 & 2.66 & 3.48 & 4.50 & 5.99 \\
DDTree \textcolor{gray!65}{(16)} & 5.13 & 7.13 & 6.13 & 8.45 & 5.39 & 7.43 & 4.91 & 6.80 & 3.09 & 4.24 & 4.93 & 6.81 \\
TreeFlash \textcolor{gray!65}{(16)} & \textbf{5.29} & \textbf{7.62} & \textbf{6.34} & \textbf{9.05} & \textbf{5.50} & \textbf{7.83} & \textbf{5.05} & \textbf{7.27} & \textbf{3.23} & \textbf{4.57} & \textbf{5.08} & \textbf{7.27} \\
\cmidrule(lr){1-13}
EAGLE3 \textcolor{gray!65}{(64)} & 2.14 & 3.73 & 2.06 & 3.63 & 2.00 & 3.47 & 1.94 & 3.38 & 1.96 & 3.41 & 2.02 & 3.53 \\
DDTree \textcolor{gray!65}{(64)} & 6.07 & 8.24 & 7.24 & 9.75 & 6.36 & 8.55 & 5.85 & 7.90 & 3.68 & 4.90 & 5.84 & 7.87 \\
TreeFlash \textcolor{gray!65}{(64)} & \textbf{6.70} & \textbf{9.39} & \textbf{7.90} & \textbf{10.99} & \textbf{6.85} & \textbf{9.51} & \textbf{6.44} & \textbf{8.96} & \textbf{3.92} & \textbf{5.38} & \textbf{6.36} & \textbf{8.85} \\
\midrule
\addlinespace[1pt]
\multicolumn{13}{@{}l}{\textbf{Target model: Qwen3 8B}} \\
\addlinespace[1pt]
EAGLE3 \textcolor{gray!65}{(16)} & 1.88 & 3.21 & 1.95 & 3.32 & 1.87 & 3.18 & 1.74 & 2.97 & 1.69 & 2.86 & 1.82 & 3.11 \\
DFlash \textcolor{gray!65}{(16)} & 4.71 & 6.33 & 5.96 & 7.88 & 4.78 & 6.34 & 4.36 & 5.87 & 2.55 & 3.32 & 4.47 & 5.95 \\
DDTree \textcolor{gray!65}{(16)} & 5.15 & 7.14 & 6.12 & 8.42 & 5.25 & 7.20 & 4.68 & 6.47 & 3.05 & 4.14 & 4.85 & 6.67 \\
TreeFlash \textcolor{gray!65}{(16)} & \textbf{5.33} & \textbf{7.71} & \textbf{6.38} & \textbf{9.20} & \textbf{5.40} & \textbf{7.74} & \textbf{5.03} & \textbf{7.24} & \textbf{3.21} & \textbf{4.53} & \textbf{5.07} & \textbf{7.29} \\
\cmidrule(lr){1-13}
EAGLE3 \textcolor{gray!65}{(64)} & 2.17 & 3.75 & 2.23 & 3.81 & 2.16 & 3.70 & 1.96 & 3.35 & 1.93 & 3.30 & 2.09 & 3.58 \\
DDTree \textcolor{gray!65}{(64)} & 6.15 & 8.25 & 7.15 & 9.56 & 6.22 & 8.35 & 5.73 & 7.68 & 3.73 & 4.89 & 5.80 & 7.75 \\
TreeFlash \textcolor{gray!65}{(64)} & \textbf{6.73} & \textbf{9.43} & \textbf{7.99} & \textbf{11.02} & \textbf{6.79} & \textbf{9.40} & \textbf{6.41} & \textbf{8.96} & \textbf{4.00} & \textbf{5.42} & \textbf{6.39} & \textbf{8.84} \\
\bottomrule
\end{tabular*}

%% file: assets/q3_30b_table.tex
\begin{tabular*}{\columnwidth}{@{\extracolsep{\fill}}l*{4}{c}}
\toprule
\multirow{2}{*}{Method} & \multicolumn{2}{c}{HumanEval} & \multicolumn{2}{c}{MBPP} \\
 & Speedup & $\tau$ & Speedup & $\tau$ \\
\midrule
\multicolumn{5}{c}{T=1 (Sampling)} \\
\midrule
DFlash \textcolor{gray!65}{(16)} & 4.69$^{0.07}$ & 6.41$^{0.06}$ & 4.98$^{0.17}$ & 6.65$^{0.20}$ \\
DDTree \textcolor{gray!65}{(16)} & 5.25$^{0.21}$ & 7.23$^{0.23}$ & 5.73$^{0.03}$ & 7.66$^{0.07}$ \\
TreeFlash \textcolor{gray!65}{(16)} & \textbf{5.74$^{0.26}$} & \textbf{7.97$^{0.30}$} & \textbf{6.03$^{0.08}$} & \textbf{8.21$^{0.10}$} \\
\cmidrule(lr){1-5}
DDTree \textcolor{gray!65}{(64)} & 6.31$^{0.03}$ & 8.61$^{0.01}$ & 6.59$^{0.07}$ & 8.84$^{0.12}$ \\
TreeFlash \textcolor{gray!65}{(64)} & \textbf{6.98$^{0.15}$} & \textbf{9.68$^{0.12}$} & \textbf{7.20$^{0.07}$} & \textbf{9.82$^{0.10}$} \\
\midrule
\multicolumn{5}{c}{T=0 (Greedy)} \\
\midrule
DFlash \textcolor{gray!65}{(16)} & 5.50 & 7.43 & 5.41 & 7.19 \\
DDTree \textcolor{gray!65}{(16)} & 6.23 & 8.43 & 5.86 & 7.87 \\
TreeFlash \textcolor{gray!65}{(16)} & \textbf{6.56} & \textbf{8.99} & \textbf{6.20} & \textbf{8.50} \\
\cmidrule(lr){1-5}
DDTree \textcolor{gray!65}{(64)} & 7.14 & 9.66 & 6.80 & 9.12 \\
TreeFlash \textcolor{gray!65}{(64)} & \textbf{7.69} & \textbf{10.57} & \textbf{7.47} & \textbf{10.22} \\
\bottomrule
\end{tabular*}

%% file: assets/abl_model_table.tex
\begin{tabular}{lcccccccc}
\toprule
 & \multicolumn{2}{c}{MATH-500} & \multicolumn{2}{c}{HumanEval} & \multicolumn{2}{c}{MT-Bench} & \multicolumn{2}{c}{Mean} \\
\cmidrule(lr){2-3} \cmidrule(lr){4-5} \cmidrule(lr){6-7} \cmidrule(lr){8-9}
$T=$ & $0$ & $1$ & $0$ & $1$ & $0$ & $1$ & $0$ & $1$ \\
\midrule
DDTree & 9.8 & 7.8$^{0.1}$ & 8.5 & 7.6$^{0.2}$ & 4.9 & 4.5$^{0.0}$ & 7.7 & 6.6$^{0.1}$ \\
w/o AR-app. & 9.6 & 7.8$^{0.1}$ & 8.4 & 7.7$^{0.1}$ & 4.9 & 4.4$^{0.0}$ & 7.6 & 6.6$^{0.1}$ \\
w/ Linear & 10.1 & 8.1$^{0.1}$ & 8.6 & 7.9$^{0.1}$ & 5.0 & 4.6$^{0.0}$ & 7.9 & 6.8$^{0.1}$ \\
w/ Frozen & 11.0 & 8.8$^{0.1}$ & 9.3 & 8.5$^{0.1}$ & 5.4 & 4.9$^{0.0}$ & 8.6 & 7.4$^{0.0}$ \\
w/ 2-prev & 11.0 & 8.8$^{0.0}$ & 9.5 & 8.6$^{0.2}$ & \textbf{5.5} & 4.9$^{0.1}$ & \textbf{8.7} & 7.4$^{0.1}$ \\
w/ CE & 10.9 & 8.8$^{0.2}$ & 9.5 & 8.6$^{0.0}$ & 5.4 & 4.9$^{0.0}$ & 8.6 & 7.4$^{0.1}$ \\
w/o Scaling & \textbf{11.0} & \textbf{8.9$^{0.1}$} & \textbf{9.6} & 8.6$^{0.1}$ & 5.3 & 4.9$^{0.0}$ & 8.6 & 7.4$^{0.0}$ \\
TreeFlash & 11.0 & 8.7$^{0.0}$ & 9.4 & \textbf{8.7$^{0.0}$} & 5.4 & \textbf{5.0$^{0.0}$} & 8.6 & \textbf{7.5$^{0.0}$} \\
\bottomrule
\end{tabular}

%% file: assets/tree_construction.tex
\begin{minipage}{0.92\linewidth}
\raggedright
\small

\emph{1. Compute AR-approximated token distributions.} \\[-1pt]
\quad $(h_{t+1}, \ldots, h_{t+\gamma})
      \gets \mathrm{DFlash}(x_{\leq t})$ \\[-1pt]

\quad \textbf{for} $i \in [\gamma]$ \textbf{do} \\[-1pt]
\quad\quad $q_{t+i} \gets \mathrm{LMHead}(h_{t+i})$ \\[-1pt]
\quad\quad $\hat{x}_{t+i}^{(1:M)}
      \gets \operatorname{TopM}(q_{t+i})$ \\[-1pt]
\quad\quad $e_{t+i}^{(m)}
      \gets \mathrm{Embd}(\hat{x}_{t+i}^{(m)}),
      \quad m \in [M]$ \\[-1pt]
\quad \textbf{end for} \\[2pt]

\quad $h'_{t+1}
      \gets h_{t+1}
      + \mathrm{SwiGLU}(\tilde{h}_{t+1} :: e_t)$ \\[-1pt]
\quad $q'_{t+1}
      \gets \mathrm{LMHead}(h'_{t+1})$ \\[2pt]

\quad \textbf{for} $i \in \{2,\ldots,\gamma\}$ \textbf{do} \\[-1pt]
\quad\quad \textbf{for} $m \in [M]$ \textbf{do} \\[-1pt]
\quad\quad\quad $h_{t+i}^{\prime(m)}
      \gets h_{t+i}
      + \mathrm{SwiGLU}(\tilde{h}_{t+i} :: e_{t+i-1}^{(m)})$ \\[-1pt]
\quad\quad\quad $q_{t+i}^{\prime(m)}
      \gets \mathrm{LMHead}(h_{t+i}^{\prime(m)})$ \\[-1pt]
\quad\quad \textbf{end for} \\[-1pt]
\quad \textbf{end for} \\[4pt]

\emph{2. Construct the draft tree with OPT-Tree selection.} \\[-1pt]
\quad $\mathcal{T} \gets \{x_t\}$ \\[-1pt]
\quad $Q \gets \operatorname{InitQueue}(q'_{t+1})$
\hfill -- candidates keyed by path probability \\[-1pt]

\quad \textbf{while} $|\mathcal{T}| < B+1$ \textbf{do} \\[-1pt]
\quad\quad $v_{t+i} \gets \operatorname{PopMax}(Q)$ \\[-1pt]
\quad\quad $\mathcal{T} \gets \mathcal{T} \cup \{v_{t+i}\}$ \\[-1pt]

\quad\quad \textbf{if} $v_{t+i} = \hat{x}_{t+i}^{(m)}$
        for some $m \in [M]$
        \textbf{then} \\[-1pt]
\quad\quad\quad $Q \gets Q \cup q_{t+i+1}^{\prime(m)}$ \\[-1pt]
\quad\quad \textbf{end if} \\[-1pt]
\quad \textbf{end while} \\[-1pt]

\quad \textbf{return} $\mathcal{T}$

\end{minipage}

%% file: references.bib
@article{chen2021evaluating,
  title = {Evaluating Large Language Models Trained on Code},
  author = {Chen, Mark and Tworek, Jerry and Jun, Heewoo and Yuan, Qiming and Pinto, Henrique Ponde de Oliveira and Kaplan, Jared and Edwards, Harri and Burda, Yuri and Joseph, Nicholas and Brockman, Greg and Ray, Alex and Puri, Raul and Krueger, Gretchen and Petrov, Michael and Khlaaf, Heidy and Sastry, Girish and Mishkin, Pamela and Chan, Brooke and Gray, Scott and Ryder, Nick and Pavlov, Mikhail and Power, Alethea and Kaiser, Lukasz and Bavarian, Mohammad and Winter, Clemens and Tillet, Philippe and Such, Felipe Petroski and Cummings, Dave and Plappert, Matthias and Chantzis, Fotios and Barnes, Elizabeth and Herbert-Voss, Ariel and Guss, William Hebgen and Nichol, Alex and Paino, Alex and Tezak, Nikolas and Tang, Jie and Babuschkin, Igor and Balaji, Suchir and Jain, Shantanu and Saunders, William and Hesse, Christopher and Carr, Andrew N. and Leike, Jan and Achiam, Josh and Misra, Vedant and Morikawa, Evan and Radford, Alec and Knight, Matthew and Brundage, Miles and Murati, Mira and Mayer, Katie and Welinder, Peter and McGrew, Bob and Amodei, Dario and McCandlish, Sam and Sutskever, Ilya and Zaremba, Wojciech},
  journal = {arXiv preprint arXiv:2107.03374},
  year = {2021},
  url = {https://arxiv.org/abs/2107.03374}
}

@article{austin2021program,
  title = {Program Synthesis with Large Language Models},
  author = {Austin, Jacob and Odena, Augustus and Nye, Maxwell and Bosma, Maarten and Michalewski, Henryk and Dohan, David and Jiang, Ellen and Cai, Carrie and Terry, Michael and Le, Quoc and Sutton, Charles},
  journal = {arXiv preprint arXiv:2108.07732},
  year = {2021},
  url = {https://arxiv.org/abs/2108.07732}
}

@article{cobbe2021training,
  title = {Training Verifiers to Solve Math Word Problems},
  author = {Cobbe, Karl and Kosaraju, Vineet and Bavarian, Mohammad and Chen, Mark and Jun, Heewoo and Kaiser, Lukasz and Plappert, Matthias and Tworek, Jerry and Hilton, Jacob and Nakano, Reiichiro and Hesse, Christopher and Schulman, John},
  journal = {arXiv preprint arXiv:2110.14168},
  year = {2021},
  url = {https://arxiv.org/abs/2110.14168}
}

@inproceedings{lightman2024lets,
  title = {Let's Verify Step by Step},
  author = {Lightman, Hunter and Kosaraju, Vineet and Burda, Yura and Edwards, Harri and Baker, Bowen and Lee, Teddy and Leike, Jan and Schulman, John and Sutskever, Ilya and Cobbe, Karl},
  booktitle = {International Conference on Learning Representations},
  year = {2024},
  url = {https://arxiv.org/abs/2305.20050}
}

@inproceedings{zheng2023judging,
  title = {Judging {LLM}-as-a-Judge with {MT}-Bench and Chatbot Arena},
  author = {Zheng, Lianmin and Chiang, Wei-Lin and Sheng, Ying and Zhuang, Siyuan and Wu, Zhanghao and Zhuang, Yonghao and Lin, Zi and Li, Zhuohan and Li, Dacheng and Xing, Eric P. and Zhang, Hao and Gonzalez, Joseph E. and Stoica, Ion},
  booktitle = {Advances in Neural Information Processing Systems},
  year = {2023},
  url = {https://proceedings.neurips.cc/paper_files/paper/2023/hash/91f18a1287b398d378ef22505bf41832-Abstract-Datasets_and_Benchmarks.html}
}

@misc{chaudhary2023codealpaca,
  title = {Code Alpaca: An Instruction-following {LLaMA} Model for Code Generation},
  author = {Chaudhary, Sahil},
  year = {2023},
  publisher = {GitHub},
  journal = {GitHub repository},
  howpublished = {\url{https://github.com/sahil280114/codealpaca}}
}

@misc{nathawani2025nemotronposttraining,
  title = {{Nemotron-Post-Training-Dataset-v2}},
  author = {Nathawani, Dhruv and Ding, Shuoyang and Lavrukhin, Vitaly and Gitman, Igor and Majumdar, Somshubra and Bakhturina, Evelina and Ginsburg, Boris and Polak Scowcroft, Jane},
  version = {2.0},
  publisher = {{NVIDIA}},
  year = {2025},
  month = aug,
  url = {https://huggingface.co/datasets/nvidia/Nemotron-Post-Training-Dataset-v2}
}

@article{yan2025qwen3,
  title = {{Qwen3} Technical Report},
  author = {Yang, An and Li, Anfeng and Yang, Baosong and Zhang, Beichen and Hui, Binyuan and Zheng, Bo and Yu, Bowen and Gao, Chang and Huang, Chengen and Lv, Chenxu and others},
  journal = {arXiv preprint arXiv:2505.09388},
  year = {2025},
  url = {https://arxiv.org/abs/2505.09388},
  doi = {10.48550/arXiv.2505.09388}
}

@inproceedings{leviathan2023fast,
  title = {Fast Inference from Transformers via Speculative Decoding},
  author = {Leviathan, Yaniv and Kalman, Matan and Matias, Yossi},
  booktitle = {Proceedings of the 40th International Conference on Machine Learning},
  year = {2023},
  series = {Proceedings of Machine Learning Research},
  publisher = {PMLR},
  url = {https://proceedings.mlr.press/v202/leviathan23a.html}
}

@inproceedings{sun2023spectr,
  title = {{SpecTr}: Fast Speculative Decoding via Optimal Transport},
  author = {Sun, Ziteng and Suresh, Ananda Theertha and Ro, Jae Hun and Beirami, Ahmad and Jain, Himanshu and Yu, Felix},
  booktitle = {Advances in Neural Information Processing Systems},
  year = {2023},
  url = {https://proceedings.neurips.cc/paper_files/paper/2023/hash/6034a661584af6c28fd97a6f23e56c0a-Abstract-Conference.html}
}

@inproceedings{miao2024specinfer,
  title = {{SpecInfer}: Accelerating Large Language Model Serving with Tree-Based Speculative Inference and Verification},
  author = {Miao, Xupeng and Oliaro, Gabriele and Zhang, Zhihao and Cheng, Xinhao and Wang, Zeyu and Zhang, Zhengxin and Wong, Rae Ying Yee and Zhu, Alan and Yang, Lijie and Shi, Xiaoxiang and Shi, Chunan and Chen, Zhuoming and Arfeen, Daiyaan and Abhyankar, Reyna and Jia, Zhihao},
  booktitle = {Proceedings of the 29th ACM International Conference on Architectural Support for Programming Languages and Operating Systems},
  year = {2024},
  publisher = {Association for Computing Machinery},
  url = {https://arxiv.org/abs/2305.09781},
  doi = {10.1145/3620666.3651335}
}

@article{cai2024medusa,
  title = {{Medusa}: Simple {LLM} Inference Acceleration Framework with Multiple Decoding Heads},
  author = {Cai, Tianle and Li, Yuhong and Geng, Zhengyang and Peng, Hongwu and Lee, Jason D. and Chen, Deming and Dao, Tri},
  journal = {arXiv preprint arXiv:2401.10774},
  year = {2024},
  url = {https://arxiv.org/abs/2401.10774}
}

@article{ankner2024hydra,
  title = {{Hydra}: Sequentially-Dependent Draft Heads for {Medusa} Decoding},
  author = {Ankner, Zachary and Parthasarathy, Rishab and Nrusimha, Aniruddha and Rinard, Christopher and Ragan-Kelley, Jonathan and Brandon, William},
  journal = {arXiv preprint arXiv:2402.05109},
  year = {2024},
  url = {https://arxiv.org/abs/2402.05109}
}

@inproceedings{li2024eagle2,
  title = {{EAGLE}-2: Faster Inference of Language Models with Dynamic Draft Trees},
  author = {Li, Yuhui and Wei, Fangyun and Zhang, Chao and Zhang, Hongyang},
  booktitle = {Proceedings of the 2024 Conference on Empirical Methods in Natural Language Processing},
  year = {2024},
  publisher = {Association for Computational Linguistics},
  pages = {7421--7432},
  url = {https://aclanthology.org/2024.emnlp-main.422/},
  doi = {10.18653/v1/2024.emnlp-main.422}
}

@article{wang2025opttree,
  title = {{OPT}-Tree: Speculative Decoding with Adaptive Draft Tree Structure},
  author = {Wang, Jikai and Su, Yi and Li, Juntao and Xia, Qingrong and Ye, Zi and Duan, Xinyu and Wang, Zhefeng and Zhang, Min},
  journal = {Transactions of the Association for Computational Linguistics},
  year = {2025},
  url = {https://direct.mit.edu/tacl/article/doi/10.1162/tacl_a_00735/128189/OPT-Tree-Speculative-Decoding-with-Adaptive-Draft},
  doi = {10.1162/tacl_a_00735}
}

@inproceedings{li2025eagle3,
  title = {{EAGLE}-3: Scaling up Inference Acceleration of Large Language Models via Training-Time Test},
  author = {Li, Yuhui and Wei, Fangyun and Zhang, Chao and Zhang, Hongyang},
  booktitle = {Advances in Neural Information Processing Systems},
  year = {2025},
  url = {https://openreview.net/forum?id=4exx1hUffq},
  note = {Poster}
}

@article{liu2025tidar,
  title = {{TiDAR}: Think in Diffusion, Talk in Autoregression},
  author = {Liu, Jingyu and Dong, Xin and Ye, Zhifan and Mehta, Rishabh and Fu, Yonggan and Singh, Vartika and Kautz, Jan and Zhang, Ce and Molchanov, Pavlo},
  journal = {arXiv preprint arXiv:2511.08923},
  year = {2025},
  url = {https://arxiv.org/abs/2511.08923}
}

@article{li2025diffuspec,
  title = {{DiffuSpec}: Unlocking Diffusion Language Models for Speculative Decoding},
  author = {Li, Guanghao and Fu, Zhihui and Fang, Min and Zhao, Qibin and Tang, Ming and Yuan, Chun and Wang, Jun},
  journal = {arXiv preprint arXiv:2510.02358},
  year = {2025},
  url = {https://arxiv.org/abs/2510.02358}
}

@article{sandler2025specdiff2,
  title = {{SpecDiff}-2: Scaling Diffusion Drafter Alignment For Faster Speculative Decoding},
  author = {Sandler, Jameson and Christopher, Jacob K. and Hartvigsen, Thomas and Fioretto, Ferdinando},
  journal = {arXiv preprint arXiv:2511.00606},
  year = {2025},
  url = {https://arxiv.org/abs/2511.00606}
}

@inproceedings{hu2026bridging,
  title = {Bridging Draft Policy Misalignment: Group Tree Optimization for Speculative Decoding},
  author = {Hu, Shijing and Li, Jingyang and Lu, Zhihui and Zhou, Pan},
  booktitle = {International Conference on Learning Representations},
  year = {2026},
  url = {https://openreview.net/forum?id=dwPdYFqVWO},
  note = {Poster}
}

@article{liu2026dart,
  title = {{DART}: Diffusion-Inspired Speculative Decoding for Fast {LLM} Inference},
  author = {Liu, Fuliang and Li, Xue and Zhao, Ketai and Gao, Yinxi and Zhou, Ziyan and Zhang, Zhonghui and Wang, Zhibin and Dou, Wanchun and Zhong, Sheng and Tian, Chen},
  journal = {arXiv preprint arXiv:2601.19278},
  year = {2026},
  url = {https://arxiv.org/abs/2601.19278}
}

@article{chen2026dflash,
  title = {{DFlash}: Block Diffusion for Flash Speculative Decoding},
  author = {Chen, Jian and Liang, Yesheng and Liu, Zhijian},
  journal = {arXiv preprint arXiv:2602.06036},
  year = {2026},
  url = {https://arxiv.org/abs/2602.06036}
}

@article{ringel2026ddtree,
  title = {Accelerating Speculative Decoding with Block Diffusion Draft Trees},
  author = {Ringel, Liran and Romano, Yaniv},
  journal = {arXiv preprint arXiv:2604.12989},
  year = {2026},
  url = {https://arxiv.org/abs/2604.12989},
  note = {Concurrent work}
}

@inproceedings{zhou2024distillspec,
  title={Distillspec: Improving speculative decoding via knowledge distillation},
  author={Zhou, Yongchao and Lyu, Kaifeng and Rawat, Ankit Singh and Menon, Aditya Krishna and Rostamizadeh, Afshin and Kumar, Sanjiv and Kagy, Jean-Fran{\c{c}}ois and Agarwal, Rishabh},
  booktitle={International Conference on Learning Representations},
  volume={2024},
  pages={32011--32050},
  url = {https://proceedings.iclr.cc/paper_files/paper/2024/hash/8766fbc68e1ed1cdef712ce273e0a363-Abstract-Conference.html},
  year={2024}
}

@inproceedings{vaswani2017attention,
  title={{Attention Is All You Need}},
  author={Vaswani, Ashish and Shazeer, Noam and Parmar, Niki and Uszkoreit, Jakob and Jones, Llion and Gomez, Aidan N. and Kaiser, {\L}ukasz and Polosukhin, Illia},
  booktitle={{Proceedings of the Conference on Neural Information Processing Systems (NeurIPS)}},
  year={2017}
}

@article{singh2025openai,
  title={Openai gpt-5 system card},
  author={Singh, Aaditya and Fry, Adam and Perelman, Adam and Tart, Adam and Ganesh, Adi and El-Kishky, Ahmed and McLaughlin, Aidan and Low, Aiden and Ostrow, AJ and Ananthram, Akhila and others},
  journal={arXiv preprint arXiv:2601.03267},
  year={2025}
}

@misc{AngelSlim2025,
    title={{AngelSlim}},
    author={Tencent AngelSlim Project Contributors},
    year={2025},
    month={6},
    url={https://github.com/Tencent/AngelSlim},
}

@inproceedings{dao2022flashattention,
  title = {FlashAttention: Fast and Memory-Efficient Exact Attention with IO-Awareness},
  author = {Dao, Tri and Fu, Daniel Y. and Ermon, Stefano and Rudra, Atri and R{\'e}, Christopher},
  booktitle = {Advances in Neural Information Processing Systems},
  year = {2022}
}

@inproceedings{lin2024awq,
  title = {{AWQ}: Activation-aware Weight Quantization for On-Device {LLM} Compression and Acceleration},
  author = {Lin, Ji and Tang, Jiaming and Tang, Haotian and Yang, Shang and Chen, Wei-Ming and Wang, Wei-Chen and Xiao, Guangxuan and Dang, Xingyu and Gan, Chuang and Han, Song},
  booktitle = {Proceedings of Machine Learning and Systems},
  year = {2024}
}

@inproceedings{kwon2023pagedattention,
  title = {Efficient Memory Management for Large Language Model Serving with PagedAttention},
  author = {Kwon, Woosuk and Li, Zhuohan and Zhuang, Siyuan and Sheng, Ying and Zheng, Lianmin and Yu, Cody Hao and Gonzalez, Joseph E. and Zhang, Hao and Stoica, Ion},
  booktitle = {Proceedings of the ACM SIGOPS 29th Symposium on Operating Systems Principles},
  year = {2023}
}

@inproceedings{loshchilovdecoupled,
  title={Decoupled Weight Decay Regularization},
  author={Loshchilov, Ilya and Hutter, Frank},
  year = { 2019 },
  booktitle={International Conference on Learning Representations}
}
